\documentclass[num-refs]{nbdt-article}

\usepackage{siunitx}
\usepackage{microtype}
\usepackage{graphicx}
\usepackage{subfigure}
\usepackage{booktabs} 
\usepackage{amsfonts}
\usepackage{mathrsfs}
\usepackage{amsmath, amssymb}
\usepackage{bm}
\usepackage{mathtools, cuted}
\usepackage[normalem]{ulem}
\usepackage{algorithm}
\usepackage{algpseudocode}
\usepackage{xcolor}

\newcommand{\be}{\begin{equation}}
\newcommand{\ee}{\end{equation}}
\newcommand{\ba}{\begin{array}}
\newcommand{\ea}{\end{array}}
\newcommand{\bea}{\begin{eqnarray}}
\newcommand{\eea}{\end{eqnarray}}
\newcommand{\balg}{\begin{align}}
\newcommand{\ealg}{\end{align}}
\newcommand{\bit}{\begin{itemize}}
\newcommand{\eit}{\end{itemize}}
\newcommand{\trm}[1]{\textrm{#1}}
\newcommand{\mbf}[1]{\mathbf{#1}}
\newcommand{\tbf}[1]{\textbf{#1}}
\newcommand{\mcl}[1]{\mathcal{#1}}
\newcommand{\mbb}[1]{\mathbb{#1}}
\newcommand{\msc}[1]{\mathscr{#1}}

\newcommand{\N}{\mcl{N}}

\newcommand{\bx}{\mbf{x}}
\newcommand{\bX}{\mbf{X}}

\newcommand{\bZ}{\mbf{Z}}
\newcommand{\bP}{\mbf{P}}
\newcommand{\bz}{\mbf{z}}
\newcommand{\bY}{\mbf{Y}}

\newcommand{\bSigma}{\bm{\Sigma}}
\newcommand{\bLambda}{\bm{\Lambda}}
\newcommand{\bmu}{\bm{\mu}}
\newcommand{\pvp}{\phi,\varphi}
\newcommand{\vph}{\varphi}

\newcommand{\bM}{\mbf{M}}
\newcommand{\bS}{\mbf{S}}

\newcommand{\nn}{\nonumber}

\papertype{Original Article}
\paperfield{Journal Section}

\title{Nonlinear Evolution via Spatially-Dependent Linear Dynamics for Electrophysiology and Calcium Data}

\author[1\authfn{1}]{Daniel Hernandez}
\author[2\authfn{1}]{Antonio K. Moretti}
\author[1]{Shreya Saxena}
\author[3\authfn{2}]{Ziqiang Wei}
\author[1]{John Cunningham}
\author[1]{Liam Paninski}

\affil[1]{Department of Statistics, Columbia University, USA}
\affil[2]{Department of Computer Science, Columbia University, USA}

\affil[3]{Janelia Research Campus, HHMI, USA; }

\corraddress{Daniel Hernandez, Google, NYC}
\corremail{dhernandezdiaz@google.com}

\fundinginfo{D.H was funded by NIH R01NS100066 and Simons Foundation 54296 grants. A. M. received funding from the NIH/NCI grant U54CA209997 and two NIH shared instrumentation grants, S10 OD012351 and S10 OD021764. S.S. was supported by the Swiss National Science Foundation P400P2\_186759 and NIH 5U19NS104649. Z.W. was funded by the Howard Hughes Medical Institute and by the Simons foundation collaboration on the global brain SCGB 542943SPI.}

\runningauthor{Hernandez et al.}

\begin{document}

\maketitle

\vspace{-5mm}
\begin{abstract}
Latent variable models have been widely applied for the analysis of time series resulting from experimental neuroscience techniques. In these datasets, observations are relatively smooth and possibly nonlinear. We present Variational Inference for Nonlinear Dynamics (VIND), a variational inference framework that is able to uncover nonlinear, smooth latent dynamics from sequential data. The framework is a direct extension of PfLDS; including a structured approximate posterior describing spatially-dependent linear dynamics, as well as an algorithm that relies on the fixed-point iteration method to achieve convergence. We apply VIND to electrophysiology, single-cell voltage and widefield imaging datasets with state-of-the-art results in reconstruction error. In single-cell voltage data, VIND finds a 5D latent space, with variables akin to those of Hodgkin-Huxley-like models. VIND's learned dynamics are further quantified by predicting future neural activity. VIND excels in this task, in some cases substantially outperforming current methods.
\vspace{-2mm}
\end{abstract}

\section{Introduction}

In recent years, advances on neural data acquisition have made it possible to record the simultaneous sequential activity of up to thousands of neurons \cite{paninski2017neural}. The analysis of these datasets often focuses on dimensionality reduction techniques that encode the activity of the population in a lower dimensional latent trajectory \cite{cunninghamyu}. At the other extreme, there is a large body of detailed electrophysiological data coming from voltage measurements in single cells \cite{allennature}. In this setting it is understood that the underlying dynamics are in fact highly nonlinear and  multidimensional, though the experimenter only has access to a one-dimensional ($1$D) observation. From such $1$D recordings, the task is to approximately recover the complete latent space paths and dynamics.

A host of sophisticated techniques has been proposed for the analysis of complex sequential data that is not well described by linear transitions and observations \cite{archer, chung, archergao, gao2014, hernandez, johnson, deepkalman, krishnan2,slds, lfads2, lfads1, park, pillow2, pillow, zhao2019streaming}. Here we present Variational Inference for Nonlinear Dynamics (VIND), a method particularly well-suited for the inference of latent dynamics from relatively smooth time series, such as those typical of neuroscience experiments. The main contribution of VIND is an algorithm that allows variational inference (VI) from structured, intractable approximations to the posterior distribution. In particular, VIND can handle variational posteriors that (i) represent nonlinear evolution in the latent space, and (ii) disentangle the latent dynamics (transition) from the data encoding (recognition). Crucially, the VIND approximate posterior shares the exact nonlinear structure of latent dynamics evolution with the model for data generation. This makes the VIND approximation potentially more powerful than models in which the choice of approximate posterior is made solely on grounds of tractability.


In this work we focus on a VIND variant in which the latent dynamics is represented as a Locally Linear Dynamical System (LLDS). The running time of LLDS/VIND is linear in the number of time points in a trial. We are especially interested in determining LLDS/VIND's ability to find hidden dynamics. After training, can the VIND-trained model generate neural data that is indistinguishable from the original observations, if provided with a suitable starting point?

In the second half of this work we apply VIND to four datasets, one synthetic and the remaining three using experimental data from multi-electrode electrophysiology, single-cell voltage recordings, and dorsal cortex calcium imaging respectively. We show that VIND excels in generative tasks, in some cases outperforming established methods by orders of magnitude in the predictive mean squared error (MSE).

\section{Background}

For a set of temporally ordered, correlated, noisy observations $\bX \equiv \{\bx_1,\dots\bx_T\}$, $\bx_t\in\mbb{R}^{d_X}$, a latent variable model proposes an additional, time-ordered set of random variables $\bZ\equiv\{\bz_1,\dots\bz_T\}$, $\bz_T\in \mbb{R}^{d_Z}$ that is hidden from view. The hidden state $\bz_t$ is endowed with a stochastic dynamics: $\bz_{t+1} \sim p(\bz_{t+1}|\bz_{1:t})$ by which it evolves. The observations $\bx_t$ are generated by drawing samples from a $\bz_t$-dependent probability distribution.

\tbf{Variational Inference.}  A naive objective for such a model is the marginal log-likelihood $\log p(\bX)$, with the latent variables integrated out of the joint. However,  marginalization with respect to $\bZ$ is intractable when the likelihood and prior are non-conjugate and the dynamics are nonlinear~\cite{bishop}. VI overcomes this problem by approximating the posterior $p(\bZ|\bX)$  with a distribution $q(\bZ|\bX)$, the  Recognition Model (RM), from a tractable class. The objective becomes the celebrated ELBO, a lower bound to $\log p(\bX) $ \cite{jordan}:
\be
\log p(\bX) \geq \msc{L}_{\trm{ELBO}}(\bX) = \underset{q}{\mbb{E}}[\log p(\bX, \bZ)] - \underset{q}{\mbb{E}} [\log q(\bZ|\bX)] \label{ELBO} \,.
\ee
Auto Encoding Variational Bayes~\cite{kingma} (AEVB) trains both 
$q_\phi(\mathbf{Z}|\mathbf{X})$ and $p_\theta(\mathbf{X},\mathbf{Z})$ simultaneously. 
Monte Carlo samples from the recognition distribution are used to compute the expectation  in Eq. (\ref{ELBO}). 

\tbf{Structured generative models.} We consider the joint density $p(\bX,\bZ)$. Our focus is on factorizations of the form:
\be
p(\bX,\bZ) \equiv p_{\phi, \theta}(\bX,\bZ) = c_{\phi,\theta}\cdot H_\phi(\bZ)  \prod_{t=0}^T f_\theta(\bx_t|\bz_t) \,, \label{mgen}
\ee
where the distribution parameters have been written explicitly. The unnormalized densities $f_\theta$ stand for an observation model that, for the purposes of this work, can be either Gaussian, $\bx_t|\bz_t \sim \N\big(m_\theta(\bz_t), \bSigma\big)$, or Poisson, $\bx_t|\bz_t \sim\trm{Poisson}\big(\lambda_\theta(\bz_t)\big)$. The respective mean, $m_\theta(\bz_t)$, and rate, $\lambda_\theta(\bz_t)$, are nonlinear functions of the latent state $\bz_t$, that we represent as neural networks. The standard deviation $\bSigma$ of the Gaussian observation model is taken to be $\bz_t$-independent. $c_{\phi, \theta}$ is a normalization constant. $H_\phi$ is the latent evolution term in $\bZ$-space with a Markov Chain structure \cite{johnson, deepkalman, krishnan2, archer, archergao}:
\begin{align}
  H_\phi(\bZ) & = h_0(\bz_0) \prod_{t=1}^T h_\phi(\bz_t | \bz_{t-1})\,, \label{H}  \\
  \bz_0 & \sim \N\big( a_0, \Gamma_0 \big) \,, \\
  \bz_t | \bz_{t-1} & \sim \mcl{N}\big( a_\phi(\bz_{t-1}),\, \Gamma \big) \,, \label{z|z}
\end{align}
where $a_\phi(\bz)$ is a nonlinear function, usually parameterized by a neural network, and $\Gamma$ is a trainable parameter. 

From Eq.~(\ref{mgen}), the posterior distribution of the Generative Model (GM) can be factorized as
\be
p_{\phi,\theta}(\bZ|\bX) = \frac{c_{\phi,\theta}\prod f_\theta(\bx_t|\bz_t)}{p_{\phi,\theta}(\bX)} \cdot H_\phi(\bZ) \,. \label{factor-posterior}
\ee

\section{Variational Inference for Nonlinear Dynamics (VIND)}

\tbf{Approximate posterior.} Successful VI relies on the choice of the approximation $q(\bZ|\bX)$. This choice is constrained by two desirable features that stand in tension: expressiveness and tractability. Specifically, we are interested in representing nonlinear flow in the latent space. Taking Eq.~(\ref{factor-posterior}) as a guidance, we therefore propose to include the GM evolution term $H_\phi(\bZ)$ into the variational posterior. That is, we first consider a posterior that factorizes as:
\be
Q_{\phi,\vph}(\bZ|\bX) = \kappa_{\phi,\vph}(\bX)\, G_\vph(\bX, \bZ) H_\phi(\bZ) \label{Q} \,.
\ee
The distinguishing feature of VIND is this reusage of the generative evolution term in the Recognition Model. 

By design, the factor $G_\vph$ in Eq.~(\ref{Q}) contains all the dependence on the observations $\bX$. For definiteness, the case
\begin{align}
  G_\vph(\bX, \bZ) & = \prod_{t=0}^T g_\vph(\bz_t|\bx_t) \,,\quad  \\
  \bz_t|\bx_t & \sim \mcl{N}(\mu_\vph(\bx_t), \sigma_\vph(\bx_t)) \,, \label{G}
\end{align}
is considered in this work, where $\mu_\vph(\bx)$ and $\sigma_\vph(\bx)$ are nonlinear maps. In Eq.~(\ref{Q}), $\kappa_{\phi, \vph}$ is a normalization constant. We note that, regardless of the specific form of $G_\vph$, $\kappa_{\phi, \vph}$ cannot be computed in closed form. In particular, the non-Gaussian term $h(\bz_T|\bz_{T-1})$, after integration with respect to $\bz_T$, yields an intractable $\bz_{T-1}$-dependent factor, see App.~\ref{app:toy-model}. As a consequence of the shared evolution, VI cannot be formulated directly in terms of $Q_{\phi,\vph}(\bZ|\bX)$.

VIND represents a way out of this conundrum that, effectively, allows for the use of an intractable, unnormalized $Q_{\phi,\vph}(\bZ|\bX)$ as the Recognition Model in VI. In what follows, we refer to $Q_{\phi,\vph}(\bZ|\bX)$ as the \emph{parent} distribution. VIND's idea is to compute a Gaussian approximation $q_{\phi, \varphi}(\bZ|\bX)$  to  the parent; this \emph{child} distribution then being used as the actual variational posterior  in Eq.~(\ref{ELBO}).
The inference problem becomes tractable since the child is normal. Importantly, the parameters in $q_{\phi, \varphi}(\bZ|\bX)$, with respect to which we optimize, are inherited from the parent. 
After training, they can be replaced back into $Q_{\phi,\varphi}(\bZ|\bX)$ obtaining, in particular, the nonlinear dynamics $a_\phi(\bz)$ for the latent space. 

Concretely, let the variational posterior $q_{\phi,\vph}$ be a Laplace approximation to $Q_{\phi,\vph}$,
\be
q_{\phi,\vph}(\bZ|\bX) = \N \big(\mbf{P}_{\phi,\vph}(\bX), \mbf{C}^{-1}_{\phi,\vph}(\bX) \big) \,. \label{Lap-approx}
\ee
The mean $\bP_{\phi,\vph}$ in Eq.~(\ref{Lap-approx}) is the solution to the following equation in $\bZ$,
\be
\frac{\partial}{\partial \bZ}\log Q_{\phi,\vph}(\bZ|\bX) 
= \mbf{0}  \,, \label{Peq} 
\ee
and the precision is given by
\begin{align}
  \left[\mbf{C}_{\phi,\vph}(\bX)\right]_{mn} & = \left. \frac{\partial^2}{\partial \bZ_m \partial \bZ_n}\log Q_{\phi,\vph}(\bZ|\bX) \right|_{\bZ = \bP_{\phi,\vph}(\bX)}  \equiv  \Big[ s_{\phi,\vph}\big( \bP_{\phi,\vph}(\bX), \bX \big) \Big]_{mn}  \,, \label{C}
\end{align}
where Eq.~(\ref{C}) defines $s_{\phi,\vph}$. The ELBO in Eq.~\eqref{ELBO} can then computed with respect to $q_{\phi,\vph}(\bZ|\bX)$ in Eq.~\eqref{Lap-approx}.

\tbf{Fixed-point iteration.} A closed form solution for Eq.~(\ref{Peq}) is not possible in general. However, for a large class of distributions, and in particular for any $Q_{\phi,\vph}$ such that $\log Q_{\phi,\vph}$ includes terms quadratic in $\bZ$, it is possible to rewrite Eq.~(\ref{Peq}) in the form
\be
\bZ = r_{\phi,\vph}(\bZ, \bX) \,, \label{fpi}
\ee
where $r_{\phi,\vph}$ is a nonlinear function that depends on the trainable parameters in $Q$, the data $\bX$ and the choice of the nonlinearities in $H$. In this form, Eq.~(\ref{fpi}) can be solved numerically by making use of the FPI method. That is, a numerical solution for Eq.~\eqref{fpi} is found by choosing an initial point $\bP^{(0)}$ and iterating
\be
  \bP^{(n)} = r_{\phi,\varphi}(\bP^{(n-1)}, \bX) \label{FPI} \,,
\ee
The VIND method assumes that this FPI converges. In practice, this assumption is guaranteed throughout training by appropriate choices of hyperparameters and network architectures (see supplementary material). In the following section, we will choose a specific form for $H$ as locally linear dynamics.

\tbf{Locally Linear Dynamics.} In the experiments conducted in this paper, the nonlinear dynamics is specified as $a_\phi(\bz) = A_\phi(\bz)\bz$, where $A_\phi(\bz)$ is a state-space dependent $d_Z\times d_Z$ matrix. We call this evolution rule, a Locally Linear Dynamical System, and the resulting inference algorithm LLDS/VIND. 
To derive the latter, consider a parent distribution distribution $Q_{\phi,\vph}$, as defined in Eq.~(\ref{Q}). The mean $\mu_\vph$ and the standard deviation $\sigma_\vph$ in Eq.~(\ref{G}) are represented as deep neural networks:
\be
\mu_\vph = \trm{NN}_{\vph_{\mu}}  (\bx_t) \,,\quad  \sigma_\vph = \trm{NN}_{\vph_{\sigma}}(\bx_t) \,.
\ee

The remaining ingredient of $Q_{\phi,\vph}$ is the shared evolution law $H_\phi$, Eq.~(\ref{H}). We write the $h_\phi$ factors that determine the latent evolution model as
\be
h_\vph(\bz_{t+1}|\bz_t) = exp\left\{ -\frac{1}{2}\big( \bz_{t+1} - A_\vph(\bz_t)\bz_t \big)^T \Gamma \big( \bz_{t+1} - A_\vph(\bz_t)\bz_t \big) \right\} \,, \label{h}
\ee
where $\Gamma$ is a constant precision matrix,  and $A_\phi(\bz_t)$ is specified as
  \be
  A_\phi(\bz_t) = \mbb{A} + \alpha \cdot B_\phi(\bz_t) \label{LLDSparam}
  \ee
  where $\mbb{A}$ is a state-space-independent linear transformation initialized to the identity, $B_\phi(\bz_t) = \trm{NN}_{\phi_B}(\bz_t)$, and $\alpha$ is a tunable hyperparameter of the model. 

  The parameter $\alpha$ in Eq.~\eqref{LLDSparam} parameterizes the deviation of the latent dynamics from linear. For $\alpha = 0$, LLDS/VIND reduces, both the statistical model and the algorithm, to GfLDS/PfLDS, \cite{archer, archergao} which provides a baseline for our model. Also, $\alpha$ quantifies how much the FPI cost deviates from a quadratic form in the variables $\bZ$, which is an important consideration for convergence analysis. The results of the paper were obtained by setting $\alpha = 10^{-2}$.

\begin{algorithm}[t]
  \caption{Learning VIND: At every epoch $\bP^{(\trm{ep})}_i$ is the numerical estimate of the hidden path corresponding to example $i$, while $\bP^{(\trm{ep})}_{\phi,\vph}(\bX_i)$ is the $\phi,\vph$-dependent posterior mean.}
  \begin{algorithmic}[1] \label{alg1}
    \Statex{Initialize $\phi,\vph, \theta,$ Nfpis $=3$} 
    \ForAll{$i$}
    \State Initialize $\bP^{(\trm{ep})}_i \leftarrow \bP^{(\trm{0})}_i$
    \EndFor

    \State $\trm{ep} \leftarrow 1;\, n \leftarrow 0$, 
    \State $\bP^{(\trm{ep})}_{\phi,\vph}(\bX_i) \leftarrow \bP^{(\trm{ep}-1)}_i , $
    \State $\mbf{C}^{(\trm{ep})}_{\phi,\vph}(\bX_i) \leftarrow s_{\phi,\vph}(\bP_i^{(\trm{ep}-1)}, \bX_i)$

    \While{not converged}
    \Statex \emph{\:\:\:\:\:\: \# Sample from $q_{\phi,\vph}(\bZ|\bX)$}
    \State $\bZ_i \sim  \mcl{N}\left(\bP^{(\trm{ep})}_{\phi,\vph}(\bX_i), \big( \mbf{C}^{(\trm{ep})}_{\phi,\vph}(\bX_i)  \big)^{-1} \right)$
    \Statex \emph{\:\:\:\:\:\: \# Perform gradient descent on} $\sum_i \msc{L}_{\trm{ELBO}}(\bX_i, \bZ_i)$
    \State ADAM update $\phi, \vph, \theta$ 
    \Statex \emph{\:\:\:\:\:\: \# Update $\bP$ and carry the FPI}
    \State $\bP^{(\trm{ep})}_i \leftarrow \bP^{(\trm{ep})}_{\phi,\vph}(\bX_i)|_{\phi,\, \vph}$ 
    \While{$n\leq$ Nfpis}
    \State $\bP^{(\trm{ep})}_i \leftarrow  r_{\phi,\varphi}(\bP^{(\trm{ep})}_i, \bX)$
    \State $ n \leftarrow n+1$
    \EndWhile
    \Statex \emph{\:\:\:\:\:\: \# Initialize next epoch}    
    \State $\trm{ep} \leftarrow \trm{ep} + 1;\, n \leftarrow 0$, 
    \State $\bP^{(\trm{ep})}_{\phi,\vph}(\bX_i) \leftarrow \bP^{(\trm{ep}-1)}_i , $
    \State $\mbf{C}^{(\trm{ep})}_{\phi,\vph}(\bX_i) \leftarrow s_{\phi,\vph}(\bP_i^{(\trm{ep}-1)}, \bX_i)$
    \EndWhile 
    
  \end{algorithmic}
\end{algorithm}

Eq.~(\ref{h}) corresponds to the stochastic dynamics of LLDS/VIND:
\be
\bz_{t+1} \sim A(\bz_t)\bz_t + \trm{noise} \,.
\ee
LLDS/VIND  has some desirable features:
\begin{enumerate}
\item The limit of linear evolution is easily taken as $A_\phi(\bz_t) \rightarrow \trm{const.}$.
\item $\max_{\bZ} |A_\phi(\bz_t) - \mbb{I}|$ is a simple measure of the smoothness of the latent trajectories. 
\end{enumerate}
Using Eqs.~(\ref{H}) and (\ref{G}), we obtain for the loglikelihood of the parent:
\begin{align}
  & \log Q_{\phi,\vph}  = \log C_{\phi,\vph} -  \frac{1}{2} \left[ (\bZ - \bM_\vph)^T \bLambda_\vph (\bZ - \bM_\vph) + \bZ^T \bS_\phi(\bZ) \bZ \right] \label{logQ}
\end{align}
where $\bM_\vph = \{\bmu_\vph(\bx_1),\dots,\bmu_\vph(\bx_T)\}$, $\bLambda_\vph$ is a block-diagonal precision matrix,

\be
\bLambda_\vph = \trm{diag}\{\sigma(\bx_1),\dots,\sigma_\vph(\bx_T)\} \,,
\ee
and $\bS_\phi(\bZ)$ is a state-space-dependent, block-tridiagonal covariance whose $d_Z\times d_Z$ blocks are given by:
\begin{align}
  \big[ S_\phi(\bZ)\big]_{t,\tau}  & = \left\{
                                     \begin{array}{cl}
                                       A_t^T \Gamma A_t & \quad\quad \trm{for } \tau = t \\
                                       -\Gamma A_t & \quad\quad \trm{for } \tau = t+1 \\
                                       -A_t^T \Gamma & \quad\quad  \trm{for } \tau = t-1 \\
                                       0 & \quad\quad \trm{otherwise}
                                     \end{array}
                                           \right.
\end{align}
Here $A_t \equiv A_\phi(\bz_{t})$.

Taking the gradients, Eq.~(\ref{Peq}), 
we obtain the LLDS/VIND FPI equation for the posterior mean, Eq.~(\ref{fpi}), 
with
\begin{align}
  r_{\pvp}(\bZ, \bX) & = \big[ \bLambda_\vph + \bS_\phi(\bZ)\big]^{-1} \cdot \bY(\bZ) \\ 
  \bY(\bZ) & = \bLambda_\vph\bM_\vph - \frac{1}{2}\bZ^T\frac{\partial \bS_\phi(\bZ)}{\partial \bZ} \bZ  \label{r} \,.
\end{align}
Note that the value of the constant $C_{\phi,\vph}$ is not required for the FPI step nor for the gradient descent step, thus intractability is evaded. The time complexity of VIND is $O(T)$. In particular the matrix $\bLambda_\vph + \bS_\phi(\bZ)$ can be inverted in linear time due to it being block-tridiagonal.

\tbf{VIND's algorithm.} VIND's complete algorithm includes two steps per epoch that are carried out in alternation, see Algorithm.~\ref{alg1}. The first step is a FPI that, for the current values of the parameters $\phi,\vph$, determines the mean and variance of a Laplace approximation to the parent. 
The second is a regular ADAM gradient descent update \cite{ADAM} with respect to the ELBO objective.
As it is customary, in order to estimate the gradients, the so called ``reparameterization trick'' is used. Samples are extracted from the child distribution $q_{\phi, \vph}$ via:
\be
\bZ_i = \bP_{\phi,\vph}(\bX_i) +  \left[\mbf{C}_{\phi,\vph}(\bX_i)\right]^{-1/2}\epsilon 
\ee
where $\epsilon$ is a standard normal sample, \cite{kingma, rezende}.
Upon convergence, the set of parameters $\phi,\vph,\theta$ that maximizes the ELBO can be plugged into $a_\phi(\bz)$, to obtain a dynamical rule  that interpolates between the different latent trajectories inferred from the data trials.

During training, the mean $\bP_{\phi,\varphi}$ represents the best current estimate of the latent trajectory. The FPI step in Eq.~(\ref{FPI}) for LLDS/VIND mixes all the components in $\bP_{\phi,\vph}$. In particular, the $t$-th component of  $\bP_{\phi,\vph}^{(n)}$ depends in general on all the time steps, both past and future, in $\bP_{\phi,\vph}^{(n-1)}$ via the inverse covariance in Eq.~(\ref{r}). At every training epoch, the best estimate for the path at a specific time point $t$ contains information from the complete data. VIND's algorithm is in this sense a smoother.

The FPI procedure gives a closed expression for $\bP_{\phi,\vph}^{(n)}$, and derivatives can be taken with respect to the model parameter. In practice we found that $n=2$ is enough for good convergence results. The normalization constant for $q_{\phi,\vph}(\bZ|\bX)$, the determinant of the precision matrix, can be computed in closed form making use of the fact that the precision matrix is block-tridiagonal~\cite{trba97}. 

\section{Relation to Previous Work}
\label{sec:previous}

The problem of inference for sequential data has been treated extensively in the literature. The GfLDS and PfLDS models introduced in \cite{archer, archergao} are particular cases of VIND in which the dynamics in the latent space is linear and time-invariant, i.e. $\bz_t | \bz_{t-1} \sim \mcl{N}\big( A \bz_{t-1},\, \mbf{Q} \big)$. In the jargon used in this paper, this corresponds to the situation in which the parent distribution is Gaussian, and therefore equal to its own Laplace approximation. Eq.~\eqref{Peq} can be solved analytically in this case and no FPI step is needed.

Several methods model the latent dynamics as a Gaussian Process (GP)~\cite{Yu:2009aa, park, anqiwu}, differing in the output and observation nonlinearities they assume. Gaussian Process Factor Analysis (GPFA) \cite{Yu:2009aa} assumes linear, time-invariant dynamics as well as a linear observation model, i.e. $\bx_t | \bz_t \sim \mcl{N}\big( C \bz_{t}+d,\, \mbf{R} \big)$, for some $C$, $d$, and $R$. In \cite{anqiwu} a GP is used instead as a mapping function with Poisson observations.

AESMC~\cite{anh2018autoencoding}, FIVO~\cite{NIPS2017_7235}, VSMC~\cite{pmlr-v84-naesseth18a} and SVO~\cite{moretti2019smoothing,moretti2019particle,moretti2020psvo} are methods for model inference and learning that maximize a lower bound to the marginal log likelihood, which is in turn approximated using Sequential Monte Carlo. The model learned by VIND is explicitly compared  to results obtained by these models in Sec.~\ref{results}.

In \cite{deepkalman}, Deep Kalman Filters (DKF) were proposed to handle variational posterior distributions that describes nonlinear evolution in the latent space. Their approximate posterior, analogous to the parent distribution in this paper, is plugged directly into the ELBO. This imposes some restrictions in the form the posterior can take - for instance, it must be Gaussian conditioned on the observations. VIND can handle factorizations of the parent distribution that are not restricted in this way, an example being LLDS/VIND, which has the form in Eq.~\eqref{Q}. VIND's ability to handle unnormalizable parent distributions is due to the fact that VIND's actual approximate posterior is always strictly normal. The same authors built upon their idea in \cite{krishnan2}, where a variational posterior was proposed that partially uses the conditional structure implied by the generative model. In this paper, a similar prescription is used by assuming that $Q_{\phi,\vph}$ and $p_{\phi,\vph}$ share exactly the same factorization for the latent evolution.

The authors of \cite{johnson} combine probabilistic graphical models with message passing in an approach based on conjugate priors. The approximate posterior distributions considered in that work are restricted by the conjugacy requirements, in particular, the evolution term must belong to the exponential family. VIND's parent distribution is not subject to this requirement. However, since VIND's actual approximate posterior is still Gaussian, it may be possible to combine the two methods into one that can handle both nonlinear evolution and discrete latent variables.

In \cite{chung}, Gaussian noise is added to the deterministic evolution rule of an RNN in the context of a variational autoencoder, termed VRNN. Similarly to LLDS/VIND, these authors share the evolution factorization between the generative model and the approximate posterior and, indeed, the only difference between the structure of their model and that of LLDS/VIND is that the evolution there is expressed as an RNN instead of as an LLDS. However, their inference algorithm only uses past data to estimate the hidden state at any given time. VIND's algorithm, based on the FPI, uses information both from the past and from the future to estimate the latent paths. In \cite{kalantari} a non-parametric approach was taken to determine the best latent dimension in an LDS. It would be interesting to apply those same methods to VIND. Finally, in \cite{lfads2, lfads1} a sophisticated, bidirectional, Deep Learning-based RNN architecture called LFADS was proposed with neuroscience applications in mind. 
For both LFADS and DKF, we found difficult to modify their code to compute the quantities that are used in this paper to evaluate the quality of training. However, given the expressive power of these works, we expect them to perform comparably to VIND in the tasks considered in the next section.

\section{Results}
\label{results}

We demonstrate the capabilities and performance of LLDS/VIND by applying it to four datasets. 
The first dataset consists of synthetically generated, 10-dimensional noisy observations on top of a 3D latent sequence whose evolution is dictated by an Euler discretization of the Lorenz system. This dataset is the simplest and cleanly illustrates VIND's ability to infer the underlying nonlinear dynamics. Secondly, VIND is applied to a multi-electrode neural recording from a mouse performing a delayed-discrimination task. LLDS/VIND is run with both Gaussian and Poisson observation models. It is found that while a Gaussian observation model is superior for the explaining the variance in the data, the Poisson model performs better when it comes to interpolation of the dynamics. 

The third dataset consists of a 1D voltage measurement from single-cell recordings. The problem in this case is not dimensionality reduction but rather to determine the nonlinear underlying dynamics (dimensionality expansion). Interestingly, the number of latent dimensions at which the accuracy of the VIND-extracted dynamics stabilizes coincides with the expectation from theoretical models of spiking neurons. Finally, we apply VIND to the difficult task of uncovering hidden dynamics in a dataset coming from dorsal cortex calcium imaging. We find that it is possible to model the data using a surprisingly low number of latent dimensions and show how to use VIND to reconstruct the dynamics of one side of the brain from the other.

Given an inferred starting point in state-space, the quality of the dynamics learned by LLDS/VIND can be ascertained by evolving the system $k$ steps into the future \emph{without} any input data. 
To clarify terminology, this is not strict prediction in the sense of pure extrapolation, since we use information about all $\bx_{t}$, both in the past and in the future, to infer the starting point. In order to avoid doubt, we use the term \emph{forward interpolate}.  Forward interpolation essentially tests the extent to which the dynamics are accurately learned.  We take VIND's capability for forward interpolation as the main measure of the fit's success. As we will show, this task remains highly challenging for simpler smoothing priors like the latent LDS, and it is one of the key strengths of VIND.

To make this analysis quantitative, we compute the $k$-step mean squared error (MSE$_k$) on test data, and its normalized version, the $R^2_k$, defined as
\begin{align}
  \trm{MSE}_k & =  \sum_{t=0}^{T-k} \left( \bx_{t+k} - \hat{\bx}_{t+k} \right)^2 \,,\quad
  R^2_k  = 1 - \frac{\trm{MSE}_k}{\sum_{t=0}^{T-k} \left( \bx_{t+k} - \bar{\bx} \right)^2}
\end{align}
where $\bar{\bx}$ is the data average for this trial and  $\hat{\bx}_{t+k}$ is the prediction at time $t+k$. The latter is obtained by i) using the full data $\bX$ to obtain the best estimate for $\bz_t$, ii) using $k$ times the  LLDS/VIND evolution equation $\bz_{t+1} = A_{\varphi}(\bz)\bz_t$, or $\bz_{t+1} = A\bz_t$ for the LDSs, to find the latent state $k$ time steps in the future, and iii) using the generative network to compute the forward-interpolated observation. Note that in particular, $k=0$ corresponds to the standard $R^2$. The more general $R^2_k$ ensures that VIND yields more than just a good autoencoder.
We will be comparing results obtained with LLDS/VIND to several models, namely, GfLDS, PfLDS, AESMC and GPFA (see Sec. \ref{sec:previous} for details).

\subsection{Lorenz system}
\begin{figure*}
  \begin{center}
    \includegraphics[width=0.95\linewidth]{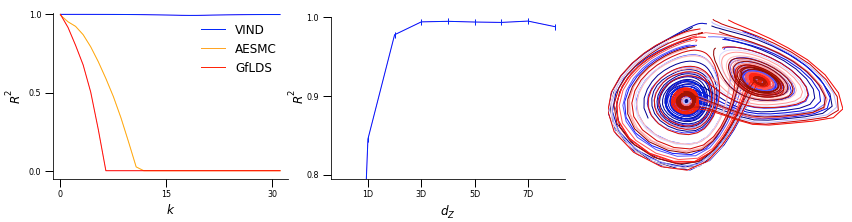}
    \caption{Comparison of results for the Lorenz dataset ($d_z = 3$) between GfLDS and VIND: (left) $R^2_k$ comparison; (center) $R^2_{10}$ as a function of dimension of the latent space; (right) VIND's inferred validation trajectories for this dataset.}
    \label{figlorenz}
  \end{center}
\end{figure*}

The Lorenz system is a classical nonlinear differential equation in 3 independent variables. 
\begin{align}
  \dot{z}_1 & = \sigma(z_2-z_1) \,, \nn \\
  \dot{z}_2 & = z_1(\rho - z_3) - z_2 \,, \\
  \dot{z}_3 & = z_1z_2 - \beta z_3\,. \nn
\end{align}
This is a well studied system with chaotic solutions that serves to cleanly demonstrate VIND's capabilities for inferring nonlinear dynamics.

We generated numerical solutions of the Lorenz system from randomly generated initial conditions, with no noise. 
Taking $\sigma=10$, $\rho=28$, $\beta=8/3$
Gaussian 10D observations were generated with the mean specified by a $\bz$-dependent neural network. The complete synthetic data consisted of 100 trials, each comprising 250 time-steps, of which 66\% was used for training and the remaining were evenly split for test and validation.

In this case, we expect the $kR^2$ to deteriorate very slowly. The results of the fit to this data are shown in Fig.~\ref{figlorenz}. The left panel shows the $R^2_k$ comparison for VIND and GfLDS fits, with $d_Z=3$. Strikingly, for this dataset, VIND's performance does not substantially deteriorate over a 30-step forward interpolation. We show in the left panel comparison with our implementation of the GfLDS and AESMC algorithms. The center panel illustrates VIND's capability to infer properties of the underlying dynamics: VIND hits peak performance at $d_Z=3$, the true dimensionality of this system. In the rightmost panel, all the paths inferred by VIND have been put together, showing the famous butterfly pattern. 

The obtained latent trajectories are topologically similar to the Lorenz attractor but do not reproduce it exactly. This is expected since VIND’s decoder can, in principle, learn to undo any smooth transformation applied to the true Lorenz trajectories. Thus, the same set of observations can be described by different sets of latent paths connected by smooth transformations.

\subsection{Electrophysiology}
\begin{figure*}
  \begin{center}
  \includegraphics[width=0.95\linewidth]{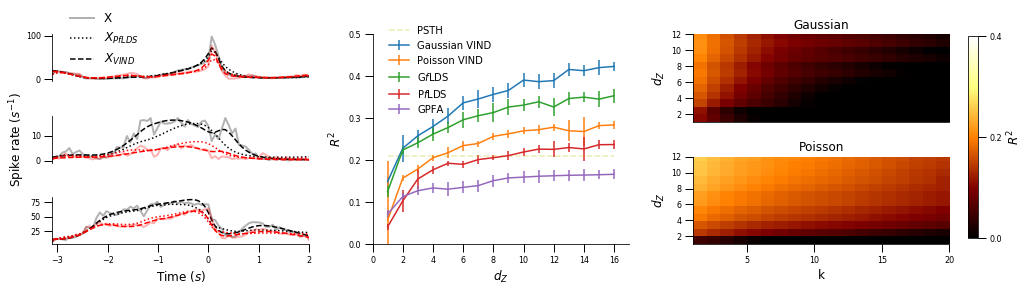}
 \caption{Electrophysiology data. (left) Sample cell spike rates, $t=0$ signals the start of the response epoch (center) Performance of explained variance ($R^2$) using different setups of VIND and other models. (right) Performance of forward interpolation ($R^2_k$) using two setups of VIND models.}
 \label{figziqiang}
 \end{center}
\end{figure*}
\label{sec:ephys}
VIND was used to analyze neural data collected from mice performing a delayed discrimination task in a simultaneous recording session (multi-unit electrophysiology) 
\cite{guoli, linature}. In this task, the animals were trained to discriminate the location of a pole using whiskers. The pole was presented at $t=-1.3$s, and an auditory go cue at $t=0$ signaled the beginning of the response epoch. 
During response, the mice reported the perceived pole position by licking one of two lick ports. Neurons in this task exhibit complex dynamics across behavioral epochs; some neurons show ramping and persistent activity from sample to delay, which relates to the preparation of the choice at response \cite{guoli, linature, ziqiang1}, while some other neurons show the peaking activity in response to the behavioral epochs, see Fig.~\ref{figziqiang}, left.

We asked whether VIND can capture the variety of neural dynamics using a few latent observations. The data was fitted for $d_Z=5$, using a Poisson observation model. 
The fit not only reproduces the neural observation, but also provides insights to the dynamics in the latent space and. Specifically, the latent paths separate cleanly by trial type, and the different epochs of the experiment can be seen.

Subsequently, a $10$-fold cross-validation method was used to decide the performance of fit using VIND's Gaussian and Poisson observation models with up to $12$ dimensions in the latent space, regardless of trial type.  The $R^2$ was computed to determine the performance of VIND as compared to other models. For VIND, both Poisson and Gaussian observation models were used. These are compared to a Peristumulus Time Histogram (PSTH), a GPFA model \cite{Yu:2009aa}, as well as GfLDS and PLDS \cite{archer, archergao}. The results are shown in the center panel in Fig.~\ref{figziqiang}. We found that nonlinear Gaussian VIND performs the best regarding explained variance of the data.

The VIND Poisson observation model gives a substantially better forward interpolation, signaling a dynamical system that more accurately represents the data evolution. This can be seen in the right panel in Fig. \ref{figziqiang}. These two results combined exemplify the VIND tradeoff between explained variance and forward interpolation capabilities. Using Poisson observations, VIND is less able to fit the higher frequency components of the data. The resulting dynamical system, however, is smoother and more appropriately captures the evolution of the system.

\subsection{Single Cell Voltage Data}
\begin{figure*}
  \begin{center}
  \includegraphics[width=0.95\linewidth]{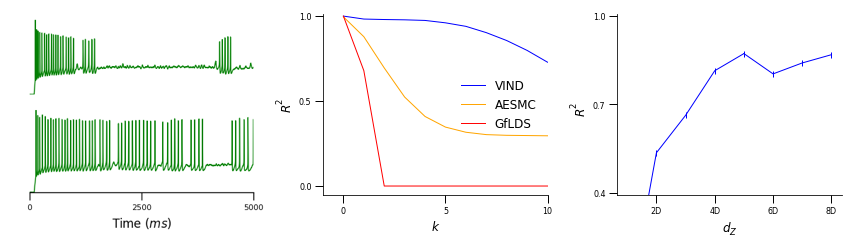}
  \caption{Summary of the LLDS/VIND fit to the Allen dataset: (left) The dataset, neurons respond to an input current; (center) VIND vs GfLDS comparison for the best 5D fits; (right) $R^2_{10}$ for different dimensions. The performance increases up to $d_Z=5$ possibly indicating the hidden dimensionality of the system.}
  \label{figallensumm}
  \end{center}
\end{figure*}

\begin{figure*}
  \includegraphics[width=\linewidth]{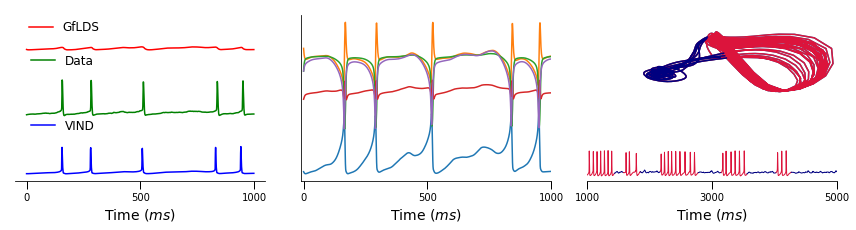}
  \caption{Inferred sample paths: (left) Original data (green) versus the 10-step (2ms) forward interpolation given by VIND and by GfLDS; (center) Latent trajectories for a 5D VIND fit of this data, showing behavior similar to the Hodgkin-Huxley gating variables; (right) A 3D cross-section of the latent space showing the representation of the spikes as big cycles (red) and the transient periods (blue).}
\label{figallenexamples}
\end{figure*}

VIND's versatility to uncover underlying dynamics is demonstrated by applying it to $1$D voltage electrophysiology data recorded from single cells. This is not a dimensionality reduction problem but rather one of recovering the latent phase space from a single variable to identify the `true dimensionality' of the system under study. The data is the publicly available Allen Brain Atlas dataset \cite{allennature}.

Intracellular voltage recordings from cells from the Primary Visual Cortex of the mouse, area layer $4$ were selected. Trials with no spikes were removed, resulting in $44$ trials from $7$ different cells. The input for each of the remaining trials consists of a step-function with an amplitude between $80$ and $151$pA. Observations were split into training ($30$ trials) and validation sets ($14$ trials). The data was then down-sampled from $50,000$ time bins (sample rate of 50 kHz) to $5,000$ in equal-time intervals, and subsequently normalized by dividing each trial by its maximal value. 

LLDS/VIND was fit to this data for $d_Z= 2,\dots,8$, repeated across $10$ runs. The top three fits were averaged and the results are summarized in Fig.~\ref{figallensumm}. The center panel displays the $R^2_{10}$ values for each choice of latent dimensionality. The fits consistently improve up to $d_Z=5$, 
after which there are diminishing returns. We note that single cell voltage data has traditionally been modeled using variants of the classical Hodgkin-Huxley neuron model (\cite{hodgkin1952quantitative}), a set of nonlinear differential equations in 4 independent variables, plus an optional independent input current. It is interesting that 5 is exactly the minimal number of latent dimensions that provide a good VIND fit for this data. The right panel displays $R^2_k$ with $d_Z = 5$ for VIND, AESMC and for GfLDS. VIND outperforms GfLDS by an order of magnitude.

The forward-interpolated observations and sample paths for selected runs of VIND and GfLDS are shown in Fig.~\ref{figallenexamples}. The left panel represents the observations over a rolling window, $k=10$ time-points in advance for both VIND and GfLDS. The dynamics inferred by GfLDS is unable to capture the nonlinear behavior in both the hyperpolarization and depolarization epochs, a task at which VIND succeeds. The VIND latent trajectories are plotted in the center panel, with the latent dimensions exhibiting similar behavior to that of Hodgkin-Huxley gating variables. In state-space, spikes are represented by big cycles (red), while interspiking fluctuations correspond to separate regions of phase space (blue). This is shown in the right panel.

\begin{figure*}
  \begin{center}
    \includegraphics[width=\linewidth]{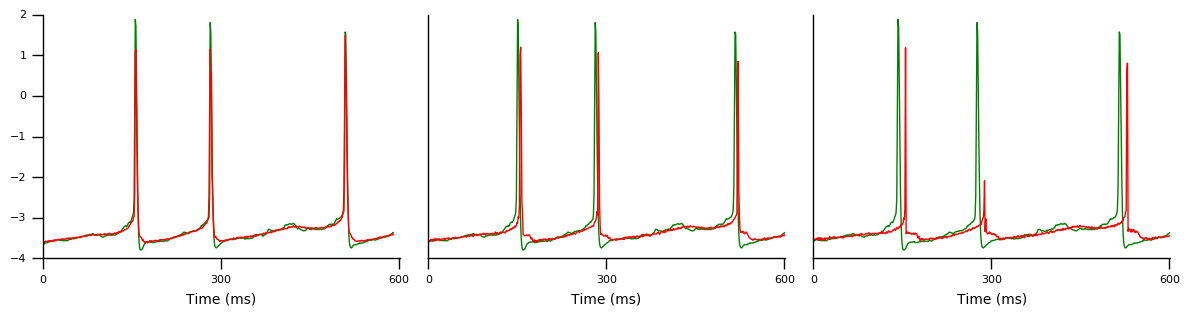}
  \end{center}
  \caption{Data (green) versus simulation of the observations (red) from the smoothed path: 10 steps ahead (left), 20 steps ahead (center), and 30 steps ahead (right). Some signs of deterioration of the prediction start to appear for the latter (failed spikes, late spiking times).}
  \label{fig7}
\end{figure*}

\begin{figure*}
  \begin{center}
    \includegraphics[width=\linewidth]{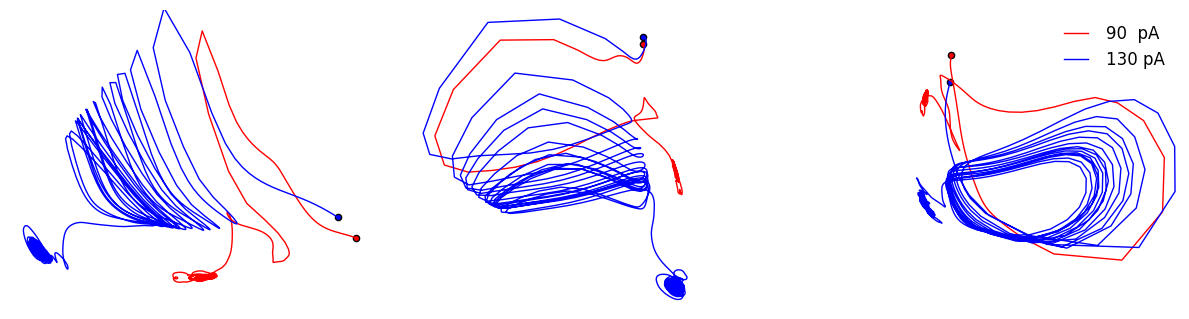}
  \end{center}
  \caption{Different views of a 3D cross section of 5D latent paths for two different trials, showing how the paths occupy different regions of state-space depending on the value of the constant input current.}
  \label{fig8}
\end{figure*}

Fig.~\ref{fig7} shows simulated paths (forward interpolation with noise) versus the corresponding real data. The expected, progressive deterioration of the VIND prediction as $k$ increases is of note. Fig.~\ref{fig8} shows several views of the same two latent paths corresponding to two different input currents showing VIND's different placement of the paths for two different input currents. 

\label{singlecell}

\subsection{Spontaneous Activity in Widefield Imaging Data}
\begin{figure*}
\centering
\includegraphics[width=\linewidth]{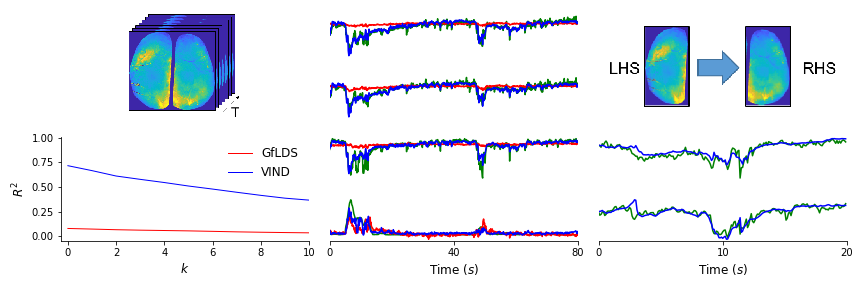}
  \vspace*{-5mm}
  \caption{Widefield Imaging Data: (top-left) An example frame of the data. The temporal dynamics and behavior signal are characterized by $X$ after preprocessing, which are simultaneously modeled using both GfLDS and VIND. (bottom-left) Variance weighted average $R^2$ values for $k$-step forward interpolation, with $d_Z=9$. (center) An example fit of $X$ using VIND on held out data. Only 4 of the signals in the $148$-dimensional $X$ signal are shown here. (right) A different VIND model was fit to the temporal dynamics of only the left hand side (LHS) of the brain ($X_{LHS}$). The latents ($Z_{LHS}$) are used to reconstruct the temporal dynamics of the right hand side (RHS) of the video ($X_{RHS}$). Fits are shown on 4 of the $66$-dimensional $X_{RHS}$ in held out data.}
\label{fig:wfom}
\end{figure*}

The unsupervised modeling of spontaneous brain activity is inherently challenging due to the lack of task structure. Here, we study the temporal dynamics of widefield optical mapping (WFOM) data and simultaneous behavior recorded from an awake head-fixed mouse during spontaneous activity \cite{ma2016wide}. This data was recorded and corrected for hemodynamics in the Laboratory for Functional Optical Imaging at Columbia University. 
An example frame of the data is shown in Fig. \ref{fig:wfom} (top-left). The preprocessing of the WFOM cortical data leads to reduced-dimension, denoised cortical activity.  Details are provided in the supplementary material. 

The temporal activity of the cortex and the movement speed (jointly called $X$) are simultaneously modeled using both GfLDS and VIND, with the results for validation data on one mouse shown in Fig. \ref{fig:wfom}, where $d_X=148$, and $d_Z=9$. The $k-$step forward interpolation is shown in Fig. \ref{fig:wfom} (bottom-left), with varying $k$, for both VIND and GfLDS. $4$ of the $148$ dimensions of $X$ and $\hat X$ on validation data are shown in Fig. \ref{fig:wfom} (center). VIND is seen to outperform GfLDS, capturing the fine-tuned dynamics in $X$, thus also leading to better interpolations. 
We highlight VIND's capability to roughly capture the dynamics of the whole superficial dorsal cortex using a $9$-D latent vector and the corresponding evolution and generative network.

Next, a VIND model was fit to the brain dynamics of only the left hand side (LHS) of the brain, after similar preprocessing of the data. Here, $d_{X_{LHS}}=60, d_{Z_{LHS}}=9$. 
A separate neural network was fit from the latents learned on the left hand side ($Z_{LHS}$) to the temporal dynamics of the right hand side (RHS) of the brain ($X_{RHS}$; $d_{X_{RHS}}=66$), with an MSE loss function. The goal was to infer dynamics from one half of the brain to the other. Fig. \ref{fig:wfom} (right) shows $5$ out of $66$ reconstructions of the temporal dynamics of the RHS in held-out data (variance weighted average $R^2 = 0.49$ for entire data). For comparison, we ran a baseline CCA analysis which yielded an $R^2$ of 0.45. This shows that the latent variables learned by VIND on one half of the brain are useful to coarsely reconstruct the temporal dynamics of the other half.

\section{Discussion}

In this work we introduced VIND, a novel variational inference framework for nonlinear latent dynamics that is able to handle intractable distributions. We successfully implemented the method for the specific case of Locally Linear Dynamical Systems, which allows for a fast inference algorithm (linear in $T$).
When applied to real data, VIND consistently outperforms other methods, in particular methods that rely on an approximate posterior representing linear dynamics and nonlinear, filtering SMC methods. Furthermore, VIND's fits yield insights about the dynamics of these systems. Highlights are the ability to identify the transition points and distinguish among trial types in the electrophysiology task, the dimensionality suggested by VIND's fits for the single-cell voltage data, and the ability of the latents learned from one half of the brain to reconstruct activity from the other half in widefield imaging data. Moreover, VIND can be naturally extended to handle labelled data and data with inputs. This is work in progress.

LLDS/VIND is written in tensorflow and the source code is publicly available.

\bibliography{sample}

\newpage

\appendix

\section{VIND's intractability}
\label{app:toy-model}
Consider a simple toy model comprising just two time steps. According to Eq.~(\ref{Q}),  $Q_{\phi,\varphi}(\mathbf{Z}|\mathbf{X})$ would be given by:
\be
Q_{\phi,\varphi}(\mathbf{Z}|\mathbf{X}) = \kappa_{\phi,\varphi}(\bX) \tilde{Q}_{\phi,\varphi}(\bZ | \bX) \,,
\ee
where $\bX = \{ \bx_1, \bx_2 \}$ and
\be
\tilde{Q}(\bZ | \bX) =  g(\mathbf{z}_0|\mathbf{x}_0) g(\mathbf{z}_1|\mathbf{x}_1) \cdot h_0(\bz_0) h(\mathbf{z}_1|\mathbf{z}_0) \,,
\ee
stands for the unnormalized distribution:
\be
\kappa_{\phi,\varphi}^{-1}(\bX) = \int \tilde{Q}(\bZ | \bX) \, d\bZ \,. \label{kappa}
\ee
Parameter subindices are suppressed in what follows for convenience.

Even in this simple setup, direct integration of $\tilde{Q}$, as in  Eq.~(\ref{kappa}), is unsuccessful. To illustrate this, consider the simplified case in which the variance parameters are all set to the identity:
\be
\Gamma_0 = \Gamma = \sigma_\varphi = \mbb{I}_{d_Z} \,.
\ee
Then, marginalizing first with respect to  $\mathbf{z}_1$:
\be
\int \tilde{Q} \, d\mathbf{z}_1 =  h(\bz_0) g(\mathbf{z}_0|\mathbf{x}_0) \cdot I(\bz_0| \bx_1)
\ee
where $I(\bz_0| \bx_1)$ is given by
\begin{align}
  I(\bz_0| \bx_1) & = \int \exp \left\{ -\frac{1}{2} \Delta( \mathbf{z}_1 | \mathbf{z}_0)^T \Delta( \mathbf{z}_1 | \mathbf{z}_0) \right. 
                   - \left. \frac{1}{2} \Delta( \mathbf{z}_1 | \mathbf{x}_1)^T  \Delta( \mathbf{z}_1 | \mathbf{x}_1) \right\} d\bz_1 \,, \label{Iz0}
\end{align}
with
\begin{align}
  \Delta( \mathbf{z}_1 | \mathbf{z}_0) & = \mathbf{z}_1 - a_\phi(\mathbf{z}_0) \,,
  \Delta( \mathbf{z}_1 | \mathbf{x}_1)  = \mathbf{z}_1 - \mu_\varphi(\mathbf{x}_1) \,.
\end{align}
Carrying out the integral,
\be
I(\bz_0| \bx_1) = \frac{1}{(2\pi)^{d_Z}} \exp\left\{ -\frac{1}{4}\big( a_\phi(\mathbf{z}_0) - \mu_\varphi(\mathbf{x}_1) \big)^2 \right\} \,.
\ee
The desired normalizing constant would then be given by
\be
\kappa^{-1} = \int  h(\bz_0) g(\mathbf{z}_0|\mathbf{x}_0) I(\bz_0| \bx_1) \, d\bz_0 \,. 
\ee
However, the argument of the exponential in the integrand includes terms in $a_\phi(\mathbf{z}_0)$ and $a_\phi(\mathbf{z}_0)^2$ which are non-quadratic in $\bz_0$. They are the source of the intractability. In turn, these are mandated by VIND's factorization of the approximate posterior, inherited from the Generative Model. 

\section{Review of the Fixed-Point Iteration method}
\label{AppFPI}

The FPI method (also known as Picard Fixed-Point Iteration) yields a numerical approximation to the solution of a system of $k$ nonlinear equations in $k$ independent variables:
\be
F_i(x) = 0 \,. \quad i = 1,\dots, k
\ee
where $x \in \mbb{R}^k$. To apply the FPI the system is transformed into the form
\be
x = T(x) \label{fpidef}
\ee
where $T:\mbb{R}^k \rightarrow \mbb{R}^k$. An initial estimate $\bx_0$ is subsequently picked. The FPI algorithm then generates the sequence $x_n$ by applying $T$ repeatedly:
\be
x_{n} = T(x_{n-1}) \,.
\ee
If this sequence converges, then it is Cauchy and its limit is the solution of Eq.~(\ref{fpidef}).

The fundamental convergence result for Picard iterations is the Picard-Banach-Cacciopoli (PBC) theorem, formulated for operators $T$, $T:X\rightarrow X$ where $(X, d_X)$ is a complete metric space:

\tbf{Theorem 1.} \emph{(PBC) Let $T$ be Lipschitz-continuous in $U\subset X$. That is}
\be
d_X\big(T(x), T(y)\big) \leq K \cdot d_X(x, y) \,, \quad \trm{ for }x,y \in U
\ee
\emph{for some real number $K$. If $K \in [0, 1)$ then $T$ has a unique fixed point $x^* \in U$ and the Picard sequence $\{x_n\}$ for $n =0, \dots, \infty$ where}
\be
x_n = T(x_{n-1}) = T^n(x_0)
\ee
\emph{converges to $x^*$ for any initial guess $x_0 \in U$.}

It can be further shown that the rate of convergence is exponential in the iteration number
\be
d_X(x_n, x^*) \leq K^n \cdot d_X(x_0, x^*) \,.
\ee
When the PBC theorem holds, we say the map $T$ is a $K$-contraction.

Let $J_{ij}(x)$ be the Jacobian of the map $T$, $i,j = 1,\dots, k$. Let $\{\lambda_i (x_0)\}$ be the eigenvalues of $J_{ij}$ evaluated at $x_0$. A common way to show that a mapping $T:\mbb{R}^k \rightarrow \mbb{R}^k$ is a contraction under the Euclidean distance in a neighborhood of $x_0 \in \mbb{R}^k$, is to show that $\max \lambda_i < 1$. In turn this can be proven using the Gershgorin Circle Theorem that gives a bound to the spectrum of a square matrix $A$:

\tbf{Theorem 2.} \emph{(Gershgorin) Let $a_{ij}$ be the entries of the square matrix $A$ and  $r_i = \sum_{j\neq i}|a_{ij}|$. Then every eigenvalue of $A$ lies within a disc centered at $a_{ii}$ with radius $r_i$.}

As a corollary, an upper bound on the maximum absolute value for the eigenvalues of $A$ is obtained:
\be
\max_i \lambda_i \leq \max_i \sum_j |a_{ij}| \,.
\ee

Applied to the Picard iteration, a sufficient condition for its convergence is obtained:
\be
\max_i \sum_j |J_{ij}| = \max_i \sum_j \left| \frac{\partial T_i}{\partial x_j} \right| < 1 \,. \label{conv-cond}
\ee
In what follows, we use this result to obtain an order-of-magnitude estimate for the VIND hyperparameters such that convergence of the VIND FPI is plausible.

\section{Implementation details of LLDS/VIND}
\label{appA}

In this appendix, we provide extra details of the VIND framework for the LLDS parameterization of the hidden dynamics.

\begin{itemize}
\item VIND is initialization-sensitive. The initial estimates for the latent path $\bP^{(0)}_i$, the starting point for the FPI, are taken to be $\bM_\vph(\bX_i)$. Moreover, empirically, we found that it is important that the initial path estimates fall within a region where the nonlinearity is not severe ($\max_{\bP_i} |A_\phi(\bz_t) - \mbb{I}| \lesssim 0.1$ for every trial $i$). This is guaranteed by proper initialization of the parameters of the recognition network.

\item The local transformation  $A_\phi(\bz_t)$ is redundant (it is akin to a gauge transformation in physics parlance). To see this, note that for every $\bz_t$, the image of the transformation $A_\phi(\bz_t) \bz_t$ is a subset of $\mbb{R}^n$. On the other hand $A_\phi(\bz_t)$ has dimensionality $\mbb{R}^{n^2}$. In other words, given $\bz_t$ and $\bz_{t+1}$, there is a continuum of matrices $A_\phi(\bz_t)$ that satisfy $\bz_{t+1} = A_\phi(\bz_t) \bz_t$. As a consequence,  $A_\phi(\bz_t)$ can be substantially restricted without loss of generality (``fixing the gauge''). In our code, $A_\phi(\bz_t)$ was constrained to be symmetric with good results.

\item The number of FPIs to produce good convergence results in Algorithm 1 (see main text) depends on the dataset. We found that $n=2$ was a good compromise that yielded convergence across datasets.

\item In all experiments, no noticeable decrease in performance was found if the gradient terms in $r_{\phi, \vph}$ - see for instance Eq.~(\ref{Y}) - and the corresponding ones for $s_{\phi, \vph}$ are neglected. These terms are subleading compared to $\bLambda_\vph \bM_\vph$ both because they are proportional $\alpha$, and because the gradient is applied on a deep neural network.
   
\end{itemize}

\tbf{FPI convergence.} As detailed in the main text, Algorithm 1, a VIND training epoch consists of two steps that are carried in alternate fashion: the FPI that updates the best estimate of the latent path, and the gradient descent step that updates the model parameters. For LLDS/VIND to converge, the FPI, defined by the map $r_{\pvp}$:
\begin{align}
  \bP & = r_{\pvp}(\bP, \bX) \\
  r_{\pvp}(\bP, \bX) & = \tilde{\bLambda}^{-1} \cdot \bY(\bP) \label{r2} \\ 
  \tilde{\bLambda} & = \bLambda + \bS(\bZ) \label{Lambdatilde} \\
  \bY(\bP) & = \bLambda_\vph\bM_\vph - \frac{1}{2}\bP^T\frac{\partial \bS_\phi(\bP)}{\partial \bP} \bP  \label{Y} \,.
\end{align}
must be in the contractive regime within a domain $D$, $D \subset \mbb{R}^{T\times d_Z}$. As remarked in App.~\ref{AppFPI}, a necessary condition for this to occur is that the Jacobian $J$ of the map $r_{\phi, \varphi}$:
\be
J_{ij}(\bZ) = \frac{\partial r_i}{\partial Z_j} \,,\quad \trm{ for } i, j \in 1,\dots,T\times d_Z \,.
\ee
satisfies Eq.~(\ref{conv-cond}).

For $\alpha = 0$, $\log Q_{\phi,\vph}(\bZ|\bX)$ is a quadratic form in $\bZ$. In this case the VIND FPI is a convex optimization problem, Eq.\eqref{Peq} is linear with a closed form solution. Therefore, deviations from convergence and convexity are always $O(\alpha)$.

In what follows, we provide an order-of-magnitude estimation to argue that $\alpha$ can be chosen to be small enough to preserve convergence and convexity in LLDS/VIND.
For the sake of clarity, we remove the parameter subindices.

For the specific case of LLDS/VIND, the entries $J_{ij}$ are suppressed both by the small hyperparameter $\alpha$ and by the gradients of the deep neural network $B_\phi(\bz_t)$, Eq.~(\ref{LLDSparam}). Neglecting the subleading terms in Eq.~(\ref{Y}) proportional to the gradient of $\bS(\bZ)$:
\be
\frac{\partial r_i}{\partial Z_j} \simeq \tilde{\bLambda}^{-1} \frac{\partial \tilde{\bLambda} }{\partial Z_j} \tilde{\bLambda}^{-1} \cdot \bLambda \bM_\vph \simeq \tilde{\bLambda}^{-1}_{ik} \frac{\partial \tilde{\bLambda}_{kl} }{\partial Z_j} \cdot r_l \,.
\ee

For an order of magnitude estimate of the necessary scales involved, let $L$ be the typical linear dimension of a bounding box in latent space inside which the latent paths are contained, 
\be
r\sim L \,.
\ee
Let $\sigma^2$ represent the typical scale of the entries of the diagonal recognition covariance matrix $\bLambda$, and let $\sigma_{\trm{ev}}^2 = \Gamma^{-1}$ represent the typical scale of the evolution covariance. Moreover, for simplicity consider the case in which $\bLambda \gtrsim \bS(\bZ)$, so that in magnitude,
\be
\tilde{\bLambda}^{-1} \sim \sigma^2\cdot\mbb{I}
\ee
Let $\Delta$ represent the typical rate of variation of the entries of the matrix $B(\mbf{z}_t)$.   Then we have
\be
\frac{\partial \tilde{\bLambda}_{kl} }{\partial Z_j} \sim \frac{\alpha \Delta}{\sigma_{\trm{ev}}^2} V_{klj}
\ee
where $V_{klj}$ is a sparse tensor (only the $(j,j)$, $(j,j+1)$ and $(j+1,j)$ blocks in $\tilde{\bLambda}_{kl}$ can depend on $Z_j$).
Replacing all these into Eq.~\eqref{conv-cond} we obtain a simple rule that, when satisfied, suggests the FPI is in the contractive regime
\be
\max_i \sum_j \left| \frac{\partial r_i}{\partial Z_j} \right| \sim c \frac{\sigma^2}{\sigma_{\trm{ev}}^2}\,\alpha \,\Delta\, L  \,.
\ee
where $c$ is an $O(1)$ constant. 

In practice, and guided by this analysis, we tune the hyperparameters and architecture of the evolution network so that
\be
\alpha \Delta \ll \frac{\sigma_{\trm{ev}}^2}{L \sigma^2}
\ee
at initialization with good results.


\section{Details of the electrophysiology task and results}
\label{appziqiang}

\begin{figure*}[!th]
  \begin{center}
    \includegraphics[width=\linewidth]{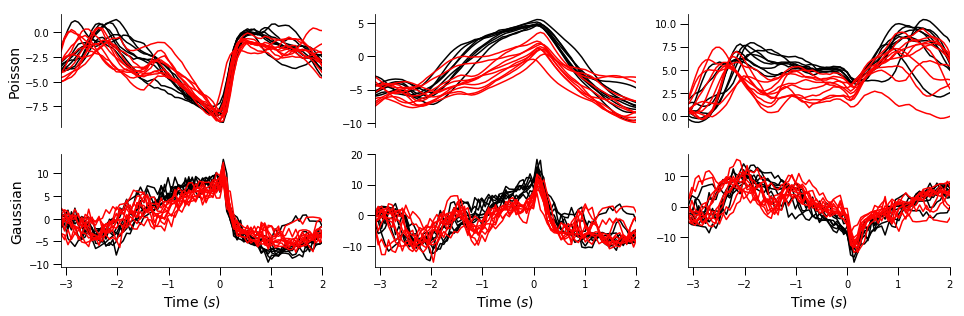}
  \end{center}
  \caption{Examples of latent dimension dynamics for Gaussian and Poisson VIND in validation data. Black lines, 
    posterior pole location; red lines, 
    anterior pole location. Notice how the inferred paths differ for posterior and anterior pole locations. Also note visible changes in dynamics at $t=-1.3$ (stimulus), and $t=0$ (go cue).}
  \label{figephyspaths}
\end{figure*}

A recording session contains 18 simultaneous recorded units, with 74 lick-left trials, (posterior pole location), 
 and 100 lick-right trials (anterior pole location). 
Spike counts were binned in a 67 ms non-overlapped time window, which resulted in a number of spike counts per bin between 0 and 10. The fit covers the time interval [-0.5, 2.0], going from the onset epoch to the end of the response epoch. At time = 0s, the mouse receives the go cue. Each trial contains 77 time bins.

Fig. \ref{figephyspaths} shows the average neuronal activity of 3 representative cells in the recordings. Cell $\#1$ is a typical neuron with small separation of trials, but strong peaking activity at transition from delay to response epochs. Cells $\#2, 3$ exhibit the stereotypical ramping activity and separations of different trial types, which are assumed for preparation of the movements. Both VIND setups (Poisson and Gaussian observations, nonlinear evolutions, $d_z=5$) can reproduce the complex and variable neural dynamics in the held-out trials (9 lick-left trials; 9 lick-right trials).  

In particular, the Gaussian VIND model can capture the changes of dynamics on finer timescales. On the other hand, the latent dynamics are smoother in the Poisson VIND model, Fig.~\ref{figephyspaths}. Smoother trajectories are correlated with superior performance in the forward interpolation tasks. Intuitively, for noisier latent paths, the algorithm attempts to ascribe some of the variance to the dynamical system, which hurts the forward interpolation capabilities. In the Poisson VIND fit represented in Fig.~\ref{figephyspaths}, the latent dynamics in dimensions 2 and 3 appears to represent the preparation of the choice where the neural dynamics for different trial types gradually diverges with time. The dynamics in latent dimension 1 shows rapid peaking dynamics at the transitions of the behavioral epochs. However, those two types of dynamics were mixed in the Gaussian VIND fit. In general, ramping and peaking dynamics is not operated by distinguishable groups of neurons, yet to our surprise they are separated in the latent space.

\section{Preprocessing of Widefield Imaging Data}
\label{sec:app_wfom}
Macro-scale wide-field optical mapping (WFOM) is an increasingly popular technique for surveying neural activity over very large areas of cortex with high temporal resolution. WFOM can image the fluorescence of genetically-encoded calcium (GCaMP6f) indicators using LED illumination and camera detection scheme. We use methods for correcting fluorescence recordings of neural activity for confounding contamination by changes in hemoglobin concentration and oxygenation as in \cite{ma2016resting}, by measuring both neural fluorescence signals and hemodynamics. This correction provides us with an accurate change in fluorescence of neural regions ($\Delta F/F$).

An example frame of the data is shown in the main text, Fig.~5, $464$-by-$473$ pixels. The activity of the mouse is simultaneously recorded using a webcam pointed at the mouse's body, and the movement speed at time $t$ is taken as a $1$D signal consisting of the standard deviation of the difference in value of all pixels from time $t-1$ to time $t$. 

We use a WFOM recording of length $2$ minutes, where the signals are sampled at $10$Hz, thus leading to $1200$ time points. We normalize $\Delta F/F$ to lie between $0$ and $1$ for every video, and then apply block singular value decomposition (SVD) to the videos for denoising and dimensionality reduction \cite{buchanan2017constrained}. First, we fit an anisotropic Wiener filter in a $4\times4$ neighborhood of each pixel to reduce uncorrelated noise while preserving spatially-local, time-correlated signals. Next, the video is partitioned into $25$ ($5\times5$) blocks, and SVD is performed on the pixels in each block. The temporal components are ranked according to a metric defined on their empirical autocorrelation function, and components that fall within a 99\% confidence interval of Gaussian white noise are discarded. Moreover, those temporal components that have a signal-to-noise ratio lower than $1.6$ are also discarded. The remaining temporal components from each block are concatenated, and these form the $X$ matrix, here $147\times1200$. This is augmented using a $1$D behavior signal that is extracted using the standard deviation of successive frames from a webcam recording the lateral view of the mouse's body, representing the speed of the mouse's movements in arbitrary units.
We used different sessions of recording from the same mouse, preprocessed in the same way, to obtain training and validation data.

\section{Details of the Sequential Monte Carlo fits}
Auto-Encoding Sequential Monte Carlo~\cite{anh2018autoencoding} is a method for model inference and learning using a variant of the ELBO constructed from the Sequential Monte Carlo marginal likelihood estimator.
In our experiments the proposal distribution factorizes into separate functions for an evolution of the latent dynamics and an encoding of the data: 
\begin{align}
Q_{SMC}(\bZ_{1:T}&|\bX_{1:T}) = \prod\limits_{t=1}^{T} h_{SMC}\big(\bz_t|\psi(\bz_{t-1}),\Gamma \big) g_{SMC}\big(\bz_t|\gamma(\bx_t),\Lambda \big) \nn
\end{align}
This choice is advantageous because $h_{SMC}(\bz_t|\bz_{t-1})$ is designed to share parameters with the evolution term of the generative model. In this way the resulting evolution term of the approximate posterior is exact.
The functions $\psi: \mathbb{R}^{d_z} \rightarrow \mathbb{R}^{d_z}$ where $\psi(\bz_t)=\bz_{t+1}$ and $\gamma: \mathbb{R}^{d_x} \rightarrow \mathbb{R}^{d_z}$ where $\gamma(\bx_t)=\bz_t$ are nonlinear time invariant represented with two layer neural networks.  We found that training separate networks for both the evolution term of the proposal and the evolution term of the generative model resulted in numerical issues when computing importance weights that caused AESMC to fail to converge.

\end{document}